\documentclass[12pt]{spieman}  
\usepackage{amsmath,amsfonts,amssymb}
\usepackage{tocloft}
\usepackage{setspace}
\usepackage{adjustbox}
\usepackage{booktabs}
\usepackage{epstopdf}
\usepackage{subcaption}
\title{Exploration of Convolutional Neural Network Architectures for Large Region Map Automation}

\author[a]{R. M. Tsenov}
\author[a,*]{C. J. Henry}
\author[b]{J. L. Storie}
\author[b]{C. D. Storie}
\author[b]{B. Murray}
\author[b]{M. Sokolov}
\affil[a]{Applied Computer Science, University of Winnipeg, 515 Portage Avenue, Winnipeg, Canada, R3B 2E9}
\affil[b]{Department of Geography, University of Winnipeg, 515 Portage Avenue, Winnipeg, Canada, R3B 2E9}

\cftpagenumbersoff{figure}
\cftpagenumbersoff{table} 
\begin{document} 
\maketitle

\begin{abstract}
Deep learning semantic segmentation algorithms have provided improved frameworks for the automated production of Land-Use and Land-Cover (LULC) maps, which significantly increases the frequency of map generation as well as consistency of production quality. In this research, a total of 28 different model variations were examined to improve the accuracy of LULC maps. The experiments were carried out using Landsat 5/7 or Landsat 8 satellite images with the North American Land Change Monitoring System labels. The performance of various CNNs and extension combinations were assessed, where VGGNet with an output stride of 4, and modified U-Net architecture provided the best results. Additional expanded analysis of the generated LULC maps was also provided. Using a deep neural network, this work achieved 92.4\% accuracy for 13 LULC classes within southern Manitoba representing a 15.8\% improvement over published results for the NALCMS. Based on the large regions of interest, higher radiometric resolution of Landsat 8 data resulted in better overall accuracies (88.04\%) compare to Landsat 5/7 (80.66\%) for 16 LULC classes. This represents an 11.44\% and 4.06\% increase in overall accuracy compared to previously published NALCMS results, including larger land area and higher number of LULC classes incorporated into the models compared to other published LULC map automation methods. 
\end{abstract}


{\noindent \footnotesize\textbf{*}C. J. Henry,  \linkable{ch.henry@uwinnipeg.ca} }

\section{Introduction}\label{section_Introduction}

\textit{Land Use and Land Cover} (LULC) maps are often the output products generated from raw satellite imagery where, in this case, each pixel of the image is assigned to a land cover class or label ({\em e.g.} water, forest, road, cropland, {\em etc.}). LULC map information is crucial for many applications such as natural resource management, wildlife habitat protection, urban expansion, hazard and damage delineation, {\em etc.} \cite{LULC:2004}. For many years, LULC maps were generated manually using semi-automated techniques such as supervised, unsupervised and hybrid algorithms ({\em e.g.}, K-means Clustering, Decision Tree and Maximum Likelihood Classier) \cite{RemoteSensingOld:2007}. However, these methods take a lot of time and effort to produce adequate accuracies but are rarely consistent over different production years or producers. Therefore, map automation is on the rise as it significantly increases the production frequency and consistency of LULC products. Methods of map automation include using advanced algorithms, machine learning tools \cite{RemoteSensingOld:2007}, or \textit{Deep Learning} (DL) \cite{DLinRemoteSensing:2019} started growing in terms of development and application in the last decade. DL has achieved success in the field of remote sensing by providing fast and consistent performance and solving problems such as image fusion, image registration, scene classification and object detection, LULC classification, {\em etc.} \cite{DLinRemoteSensing:2019}. Research on DL for remote sensing has resulted in groundbreaking papers published every few months in the fields of image processing \cite{ImageProcessing:2016}, computer vision \cite{ComputerVision:2016}, and object detection \cite{ObjectRecnog:2017}. Most issues in DL map applications are solved using Convolutional Neural Networks (CNN) as CNNs are considered the state-of-the-art method for computer vision tasks, such as segmentation, image classification and object recognition \cite{DLinRemoteSensing:2019,CNN_LitReview}.

To resolve the semantic segmentation task of LULC mapping, Long {\em et al.}\cite{FCN:2014} and Badrinarayanan{\em et al.}\cite{SegNet:2016} introduced an encoder-decoder architecture, which extracts features from the imagery using CNN in the form of an encoder and then decodes those features back to the original size using a reversed CNN. The implementation of \textit{encoder-decoder} networks in remote sensing \cite{DLLulcFirst:2015,DLLulcSecond:2015,StorieDeepLN:2018,HenryDeepLN:2018,AlhassanDeepLN2020}, has achieved great success with accurate and consistent results between 80\% and 90\% for LULC mapping. These successes have led to the implementation of DL models for mapping of LULC within commercial software, for example, ArcPro (Esri). However, in comparison to the published literature discussed above, commercially available DL models have lower overall classification accuracies, typically ranging from 78-88\% \cite{arcgis:2020} and have even lower performance with a larger number of LULC classes.

The goal of this work was to develop and improve the results of land cover mapping using DL techniques based on label or training data from the North American Land Change Monitoring System (NALCMS)\footnote{\url{http://www.cec.org/north-american-land-change-monitoring-system/}}. The NALCMS labels were chosen for training because of the large number of potential land cover categories (20 classes), ranging from arctic to tropical ecosystems, and because the NALCMS provides a much larger land area extent than other labelled/training datasets. The 30m NALCMS maps are currently only generated every five years, by the three-country consortium, due to cloud cover, technical and resource constraints \cite{latifovic:2016}. Other users are motivated to increase the frequency of production of the NALCMS products at accuracies that can only be obtained by developing automated DL approaches. For example, the overall classification accuracy of 82\% has been achieved for Mexico using MODIS data \cite{Colditz:2010} and 76.6\% accuracy for all of North America \cite{latifovic:2012}using Landsat \cite{latifovic:2016} both using 30m NALCMS.

To achieve the goal of improving overall accuracy, this work incorporated new and modified CNNs \cite{UNet:2015,ResnetV2:2016,Xception:2017,FeedbackwardDecoding:2019} to the existing architecture, and tested a variety of more advanced state-of-the-art model extensions and their combinations \cite{ProgressiveGAN:2018,Deeplabv3+:2018}. We experimented with CNNs on two different datasets, one that uses 8-bit data from Landsat 5/7, and the second uses 16-bit data from Landsat 8, both with a spatial resolution of 30m x 30m. Data from these two sensors have similar spectral resolutions with bands in the red, green, blue (RGB) and near- and shortwave-infrared regions. After determining the best model variation on the smaller datasets, two models were trained on the larger Landsat 5/7 Prairie dataset and the Landsat 8 Lake Winnipeg dataset. Landsat 5/7 Prairie dataset model resulted in a 80.66\% overall accuracy (OA), which is equivalent to the performance of ArcGIS/Pro deep learning package \cite{arcgis:2020}, while Landsat 8 Lake Winnipeg dataset model performance reached 88.04\% OA. The Landsat 8 data increased the OA by 11.44\% compared to \cite{latifovic:2016} and is ~6-8\%. Producing models for both Landsat 5/8 and Landsat 8 data means that historical and current LULC maps can be generated back to 1984 with more frequency, consistency and accuracy than ever before.
\section{Use of DL in Semantic Segmentation and Remote Sensing}\label{section_LiteratureReview}

\subsection{Deep Learning}
In the last two decades, the use of Neural Networks (NN) and DL has grown due to a large number of advances in many research areas \cite{DL_Field:2019} such as large datasets, improvements in the Graphics Processing Unit (GPU) technology, and models, which previously took weeks or months to train, now take just a few hours or days. In 2012, the AlexNet \cite{AlexNet:2012} model was presented and won the ImageNet Large Scale Visual Recognition Challenge. It was considered a revolutionary model that first showed NNs could outperform other algorithms in image recognition and classification tasks. This new architecture was only able to be trained due to the advent of general purpose computing using GPU. Advances in CNN architecture occurred in 2014, the VGGNet model \cite{VGG:2015} and GoogleNet, also known as the Inception model \cite{InceptionV1:2014}. Recently, Francois Chollet introduced an extension of the Inception architecture called Xception \cite{Xception:2017}, and He \cite{Resnet:2015} introduced groundbreaking architecture that resembles VGGNet, but is approximately eight times deeper.

CNNs were first successful at the recognition of handwritten digits using LeNet developed by LeCun in 1998 \cite{LeNet:1998} using CPUs. DL using GPUs advanced applications in speech recognition \cite{SpeechRecognition:2017}, natural language processing (NLP) \cite{NLP:2017}, image processing \cite{ImageProcessing:2016}, computer vision \cite{ComputerVision:2016}, object detection \cite{ObjectRecnog:2017}. Remote sensing applications was a natural extension of image and object recognition \cite{StorieDeepLN:2018,HenryDeepLN:2018,DLinRemoteSensing:2019,AlhassanDeepLN2020} especially in fields like image preprocessing and classification. This work focuses on mapping of LULC, within the branch of classification in DL for remote sensing.

\begin{table}
        \centering
        \caption{Comparison of popular remote sensing datasets and their state-of-the-art DL models, where most of the datasets are used as benchmarks. This table includes metrics for comparison such as Overall Accuracy (OA), mean Intersection over Union (mIoU), and F1-Score.}
        \label{tab_dataset_comparison}
        \begin{adjustbox}{width=\columnwidth,center}
        \begin{tabular}{ccccccc}
        \toprule
        Dataset & Model & Classes & OA & F1-Score & mIoU & Reference  \\

        \midrule
        INRIA & ICT-Net & 1 &               97.14 &     - & 80.32 & \cite{chatterjee2019semantic}\\
    
        AIRIS & ICT-Net & 1 &               - &         95.7 & 91.7 & \cite{chatterjee2019semantic}\\
    
        LandCover.ai & ResNet-50 & 4 &      85.56 &     - & - & \cite{Boguszewski_2021_CVPR}\\
    
        ISPRS Potsdam & ResUNet-a & 5 &     91.5 &         92.9 & - & \cite{diakogiannis2020resunet}\\
    
        ISPRS Vaihingen & TreeUNet & 5 &    90.4 &      89.3 & - & \cite{yue2019treeunet}\\
    
        GID-C & Dual Attention Deep Fusion  & 5 &       85.49 &     85.37 & 74.45 & \cite{li2021dual}\\
    
        DeepGlobe & Deep Aggregation Net & 6 &              - &         - & 52.72 & \cite{Kuo_2018_CVPR_Workshops}\\
    
        GID-F & Dual Attention Deep Fusion  & 15 &      83.93 &     83.23 & 71.29 & \cite{li2021dual}\\
    
        CLC 2018 & U-Net & 15 &                             82.41 &     77.27 & - & \cite{arcgis:2020}\\
    
        NLCD 2016 & U-Net & 16 &                            78.1 &      67.69 & - & \cite{arcgis:2020}\\
    
        GeoManitoba & ResNet-FCN & 19 &     90.46 &     - & 75.66 & \cite{VictorAutomatedLULC:2019}\\
        \bottomrule
        \end{tabular}
    \end{adjustbox}
\end{table}

\subsection{Deep Convolutional Encoder-Decoders}\label{encoder_decoder_subsection}
Many computer vision tasks cannot be solved by conventional CNNs, such as object detection or semantic segmentation. However, with architectural structural changes to the networks, they can be used in these domains. To solve these tasks, the deep convolutional encoder-decoder architecture was first introduced in \cite{FCN:2014}. The concept of the encoder-decoder is to extract features from the input using a CNN and then decode those features back to the original size using a modified CNN before classification. This concept was later improved in \cite{UNet:2015,SegNet:2016} by slightly altering existing architecture and introducing more layers to the decoder. At the same time, Deeplab introduced in \cite{Deeplabv1:2016,Deeplabv3+:2018} focused more on modifying the encoder with the CNNs such as ResNet\cite{ResnetV2:2016}, and Xception \cite{Xception:2017}. Similarly, DeepLab expanded the architecture with network extensions, such as Atrous Spatial Pyramid Pooling (ASPP) and Output Stride (OS) to capture objects and context of the features at multiple scales. At the same time, researchers starting focusing on areas other than end-to-end network architectures. For instance, some researchers focused on network extensions, which are plug-in modules that aim to improve the performance of the deep convolutional encoder-decoders without significant changes to the existing architecture. In 2015, the context aggregation extension module \cite{Context:2016} was presented to increase the sharpness of predictions by aggregating multi-scale contextual information. Additionally, Goodfellow introduced a Generative Adversarial Network (GAN) \cite{GAN:2014,SemanticGAN:2016}. GAN consist of a generator and a discriminator, where the goal of the generator is to produce an image as close to the real one as possible, and the discriminator attempts to distinguish between original and generated images. \textit{Progressive GAN}, introduced in 2017 \cite{ProgressiveGAN:2018,SemanticProgressiveGAN:2019} improved the original GAN by gradually adding layers to both generator and discriminator throughout the training, which helps with the training stability.

\subsection{CNNs in Remote Sensing}
The success of DL in image processing encouraged researchers to use it in remote sensing. The development of CNNs was mainly focused on scene classification and LULC map production. One of the first studies classified LULC using high-resolution satellite images (UC Merced dataset) \cite{DLLulcFirst:2015,DLLulcSecond:2015} due to good structural information. At that time, conventional DL algorithms performed poorly on medium-resolution satellite images (10m-30m per pixels) because of a deficiency of such structural information. In 2017, Sharma presented a patch-based CNN \cite{patch-basedCNN:2017}, which performed effectively on medium-resolution satellite images. Also, Multi-Size/Scale ResNet Ensemble (MSRE) architecture was invented by \cite{latifovic:2019,Pouliot:2021} to perform a single-pixel prediction based on an input image cropped to a specific size.

Most of the modern architectures follow the encoder-decoder structure with some minor changes. One of the first to successfully use semantic segmentation architectures for remote sensing applications were \cite{StorieDeepLN:2018,HenryDeepLN:2018,VictorAutomatedLULC:2019,AlhassanDeepLN2020}. Their solution is based on an FCN structure with different encoders ({\em i.e.} VGGNet, GoogleNet, ResNet) used on Landsat 5/7 satellite images. Also, they implemented GAN and context module extensions to boost the performance of predictions. Additionally, work presented in 2018 \cite{ModifyCNNforLULC:2018} showed that architectures that were successful in semantic segmentation on the real-world images need to be modified to get their full potential in remote sensing. Many different modifications were made to improve the performance of the architectures on satellite imagery. For example, to keep most of the structural information, a no-downsampling encoder network that relies on atrous convolution was developed \cite{NoDownsmapling:2016}. Some works centred on modifying the decoder by adding symmetric unconvoluted layers \cite{unconvolutedCNN:2018,unconvolutedCNN2:2018}. These architectures significantly improved the accuracy of generated LULC maps.


In the field of remote sensing, there are a number of datasets and state-of-the-art DL models that focus on resolving object detection and semantic segmentation problems. Table \ref{tab_dataset_comparison} compares the most popular datasets and DL algorithms from literature and software. One trend that can be seen is that the lower the number of classes, the better the performance of the models. When the number of LULC classes exceeds 10, only ResNet-FCN was shown to obtain an OA higher than 85\%. Also, other models that provide accurate results follow the U-Net architecture \cite{UNet:2015}, Faster R-CNN \cite{FasterRCNN:2015}, and Mask R-CNN \cite{MaskRCNN:2017}. The last two derive from R-CNN \cite{RCNN:2014} architecture that relies on splitting images into segments then processing each by a DCNN.

\begin{figure}
    \centering
    \includegraphics[width=0.75\linewidth]{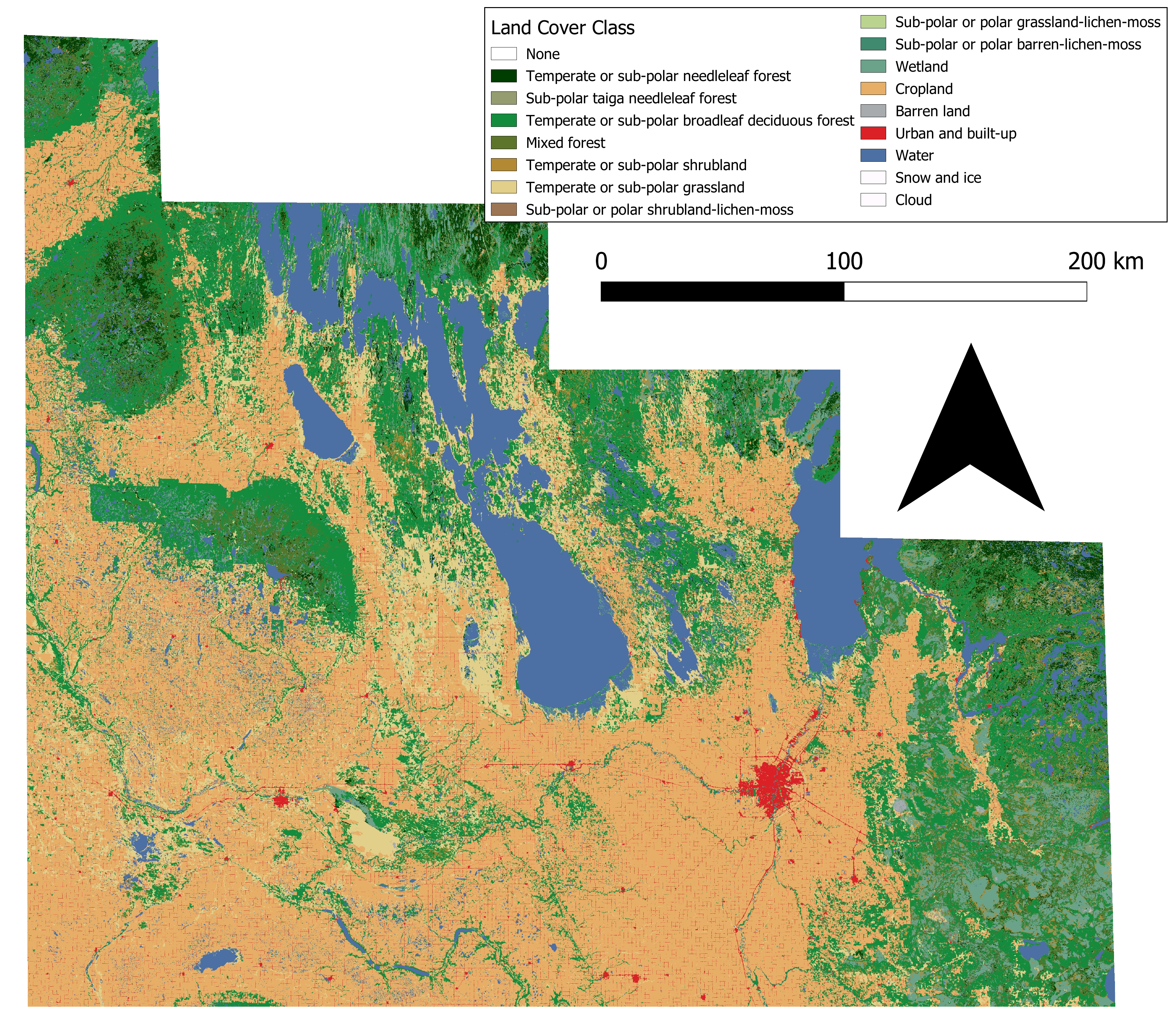} 
    \caption{Southern Manitoba land cover based on NALCMS labels and Landsat 5/7 data.}
    \label{fig:Small_Dataset}
\end{figure}

\begin{figure}
    \centering
    \includegraphics[width=1\linewidth]{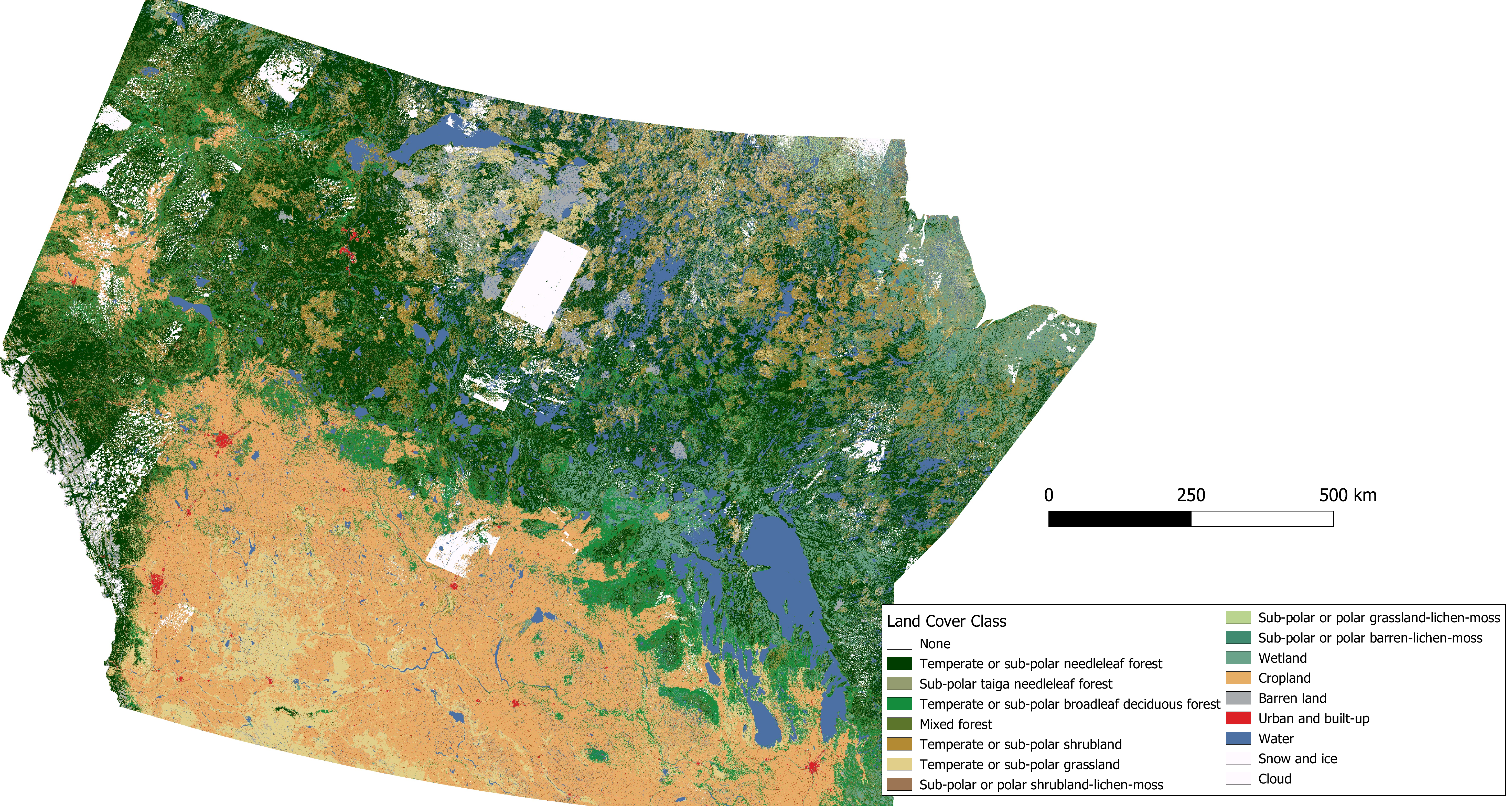} 
    \caption{Prairie land cover based on NALCMS labels and Landsat 5/7 data.}
    \label{fig:Big_Dataset}
\end{figure}

This work is a continuation of \cite{VictorAutomatedLULC:2019}, which was able to reach OA higher than 90\% on the dataset with 19 classes. While these results are good, the work was conducted on a relatively small geographical area ($\sim$ 148,800 $km^2$). In this paper, Landsat 5/7 data was used for mapping the geographic extent of both southern Manitoba (Figure \ref{fig:Small_Dataset}) and the Prairies (Figure \ref{fig:Big_Dataset}), while Landsat 8 was used for the entire Lake Winnipeg/Churchill-Nelson watershed (not shown).

\section{Implementing and Training the Models}\label{section_Implementations}

\subsection{Data}
As the overall goal of this work was to develop models to automate LULC map production corresponding to the NALCMS labels; the models were developed using both the 2010 NALCMS labels with Landsat 5/7 data (2009-2011), as well as with 2015 NALCMS labels with Landsat 8 (2014-2016) data. In the Prairie (Figure \ref{fig:Big_Dataset}) and Lake Winnipeg/Churchill-Nelson watershed, 15 of the 19 NALCMS classes were present and a cloud label was added to make it 16 classes. Cloud labels were generated using Otsu thresholding techniques on the blue band \cite{Otsu:1979} for the Landsat 5/7 dataset; cloud identification was done using the Quality Assessment (QA) band for the Landsat 8 dataset.

Due to the associated training time required for an area the size of the watershed dataset, to allow for direct comparison to \cite{VictorAutomatedLULC:2019}, and to expand to larger regions of interest, we used three datasets in our analysis. The dataset were: (i) the southern Manitoba (148,800 $km^2$) and (ii) the three prairie provinces of Alberta, Saskatchewan and Manitoba ($\sim$ 1,960, 000 $km^2$) based on 2010 Landsat 5/7, and (iii) the Lake Winnipeg/ Churchill-Nelson watershed (most of the prairies, plus parts of eastern Ontario and North Dakota) using 2015 Landsat 8 ($\sim$ 1,400,000 $km^2$). Landsat 5/7 data was obtained for the southern Manitoba dataset (Figure \ref{fig:Small_Dataset}), which is the same data and area coverage used in \cite{VictorAutomatedLULC:2019} to see the impact of model-architecture experiments on results. Most of the experiments were performed on the Landsat 5/7 southern Manitoba dataset to reduce the training time required for the comparison of models and model extensions. Then, based on the results, the best architecture was chosen, and the models for Landsat 5/7 (Prairies) and Landsat 8 (Lake Winnipeg watershed) were generated.

\subsection{Data Preprocessing}
All satellite data and corresponding labels were preprocessed before training. A popular input-size for encoder-decoder networks is 224 x 224 pixel rows and columns, so the first step was to tile the larger dataset extents. For example, the mosaic forming the southern Manitoba dataset has 14,975 rows x 13,331 columns of pixels. By creating tiles of 224 x 224 pixels, there is a suitable number of images to train a DL model, and images of this size can be processed by our computer hardware. Consequently, we used the data augmentation technique called \textit{tiling}, which was used in \cite{StorieDeepLN:2018,HenryDeepLN:2018,VictorAutomatedLULC:2019,AlhassanDeepLN2020}, with some minor improvements.

\subsubsection{Tiling}
\begin{figure}
    \centering
    \begin{subfigure}[b]{0.3\textwidth}
        \includegraphics[width=1\textwidth]{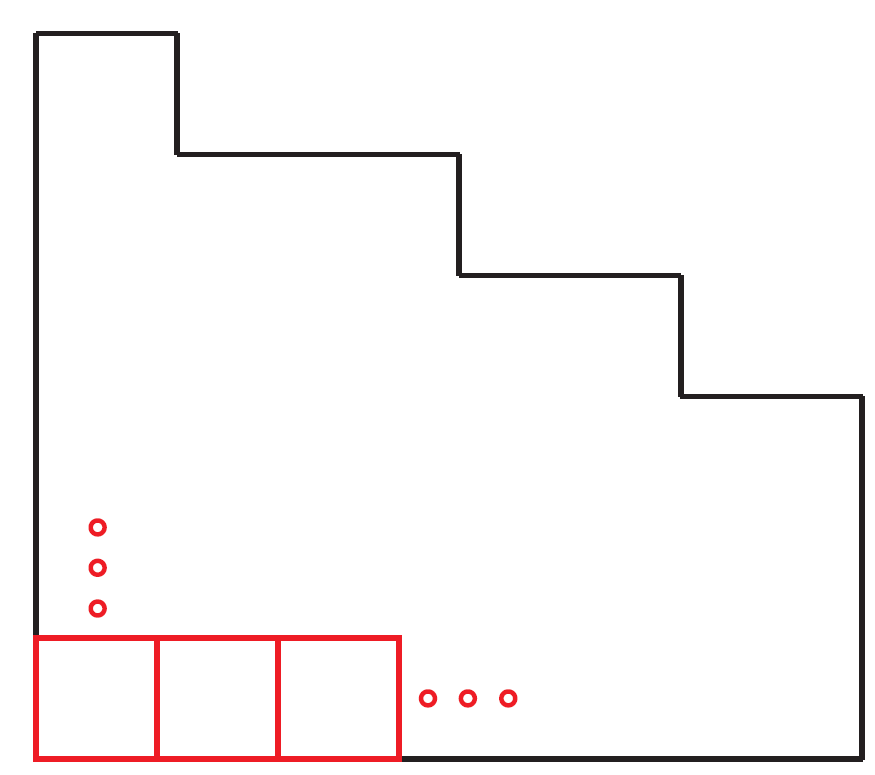}
    \end{subfigure}
    \hspace{0.02\textwidth}
    \begin{subfigure}[b]{0.3\textwidth}
        \includegraphics[width=1\textwidth]{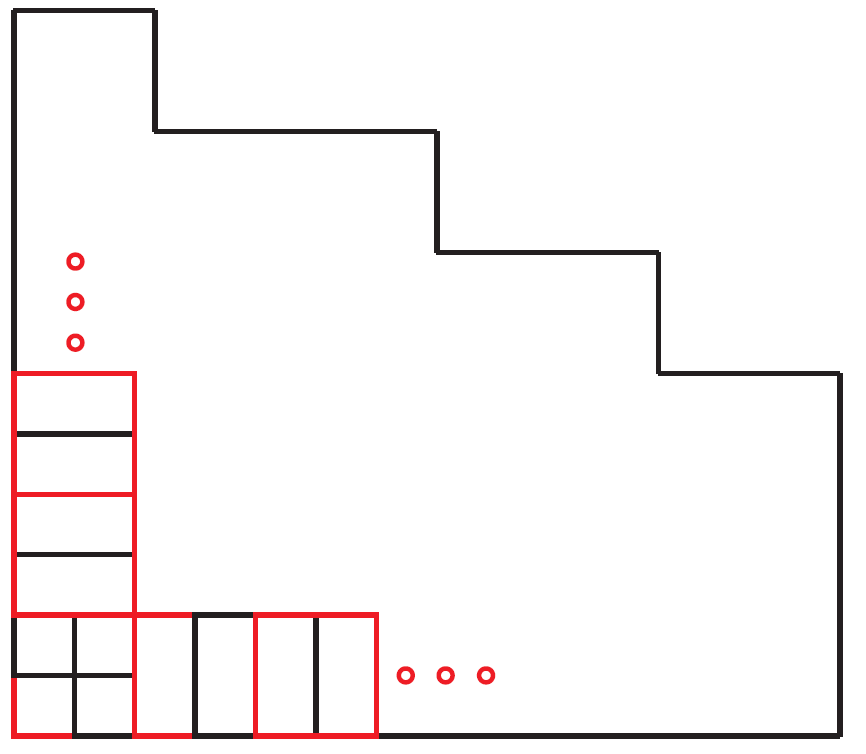}
    \end{subfigure}
    \caption{Illustration of the tilling process\cite{VictorAutomatedLULC:2019}.}
    \label{fig:Preprocess_augmentation}
\end{figure}

Tilling is the process of splitting a larger image into smaller-sized, square images (Figure \ref{fig:Preprocess_augmentation}), also referred to as tiles, where a tile is one 224 x 224 image (pixel row and column). The southern Manitoba dataset (14,975 x 13,331 pixels) generated 7,270 disjoint tiles. Moreover, increasing the number of tiles generated from the same region of interest extent can be used to improve feature detection. For example, tilling with 1/2 overlap (shifted tilling) was used to increase the size of the dataset by a factor of 4 (Figure \ref{fig:Preprocess_augmentation}), generating a dataset of 29,100 training tiles and 2,100 validation tiles. Additionally, the Landsat 5/7 Prairie dataset was generated using regular tiling and consisted of 85,400 training tiles and 8,500 validation tiles, while the Landsat 8 Lake Winnipeg watershed dataset was generated using shifted tiling and consisted of 107,900 training tiles and 12,000 validation tiles. Splitting tiles for training and validation was performed using a stratified random process to ensure that training and validation sets contain all LULC classes and tiles from all locations. The splitting ratio was 90\% training to 10\% validation, following \cite{StorieDeepLN:2018,HenryDeepLN:2018,VictorAutomatedLULC:2019,AlhassanDeepLN2020}. This splitting ratio allows effective training of the LULC classes with low presence or representation.

\subsubsection{Transformation}
\begin{figure}
    \centering
    \begin{subfigure}{0.3\textwidth}
        \includegraphics[width=0.9\linewidth]{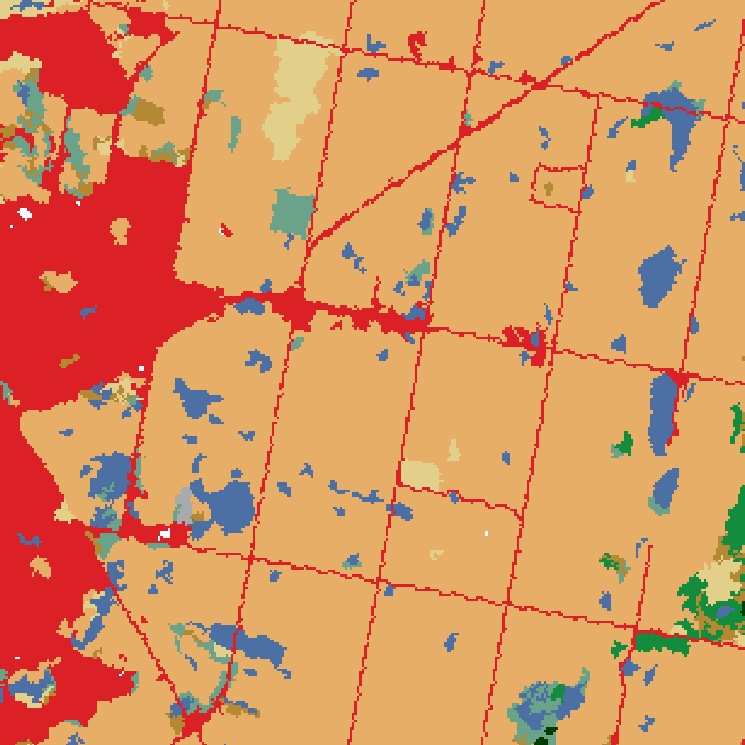} 
    \end{subfigure}
    \begin{subfigure}{0.3\textwidth}
        \includegraphics[width=1\linewidth]{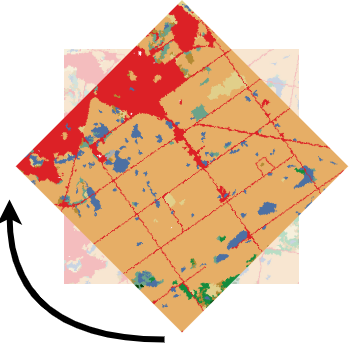}
    \end{subfigure}
    \begin{subfigure}{0.3\textwidth}
        \includegraphics[width=1\linewidth]{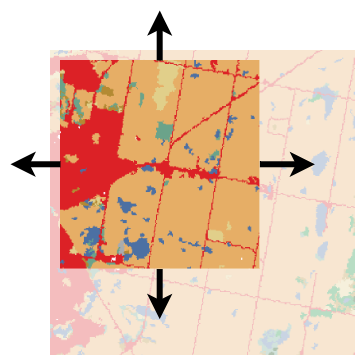}
    \end{subfigure}
    
    \begin{subfigure}{0.6\textwidth}
        \includegraphics[width=1\linewidth]{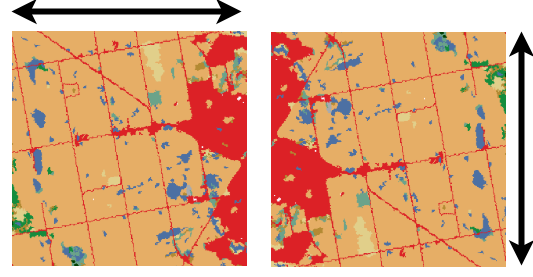}
    \end{subfigure}
    
    \caption{Figure depicting the tile transformations used during training. This includes rotating, zoom cropping, and flipping both vertically and horizontally.}
    \label{fig:Preprocess_Transformation}
\end{figure}

Training DL models often results in either overfitting or underfitting \cite{OverfitUnderfit:2019} that is caused by a lack of variation in the dataset or inadequate model architecture. One of the simplest ways to resolve this issue is to shuffle data after every epoch and apply random data transformation on each image \cite{DataTransform1:2019,DataTransform2:2019}. Data transformation ensures that tiles are not represented in the same way during training by randomly applying rotation, flipping, or zoom cropping (Figure \ref{fig:Preprocess_Transformation}). Note that, unlike augmentation (shifted tiling), transformation is used during training and does not increase the number of generated tiles.

\subsection{Experimental Design}
Training of the networks was performed using the \textit{TensorFlow} \cite{abadi2016tensorflow} Python library with \textit{Docker} containers\footnote{\url{https://www.tensorflow.org/install/docker}}. Most of the results were generated using an NVIDIA Digits DevBox\footnote{\url{https://developer.nvidia.com/devbox}} containing four Titan X GPUs with 12GB of memory per GPU, 64 GB DDR4 RAM, and a Core i7-5930K 3.5 GHz processor. The training time on the southern Manitoba datasets varied between 5 and 7 days, while the Prairie and Lake Winnipeg watershed dataset took between 20 and 30 days. The prediction time of the network on the non-augmented southern Manitoba dataset took 8 minutes, while the prediction of the Prairie or Lake Winnipeg watershed datasets took close to 1.5 hours for both sensors. Most of the networks were trained with a batch size of 16 on a model that was distributed across 3 GPUs, leading to a global batch size of 48. However, some network variations used a lower batch size due to computational complexities. As per \cite{VictorAutomatedLULC:2019}, training was performed with a learning rate of $\eta=10^{-4}$ for 100 epochs, then with a learning rate of $\eta=10^{-5}$ for another 100 epochs using the Adam optimization algorithm \cite{Adam:2017}.

\section{Experiments, Results, and Analysis}\label{section_Results}

\begin{figure}
	\centering
	\includegraphics[width=0.85\linewidth]{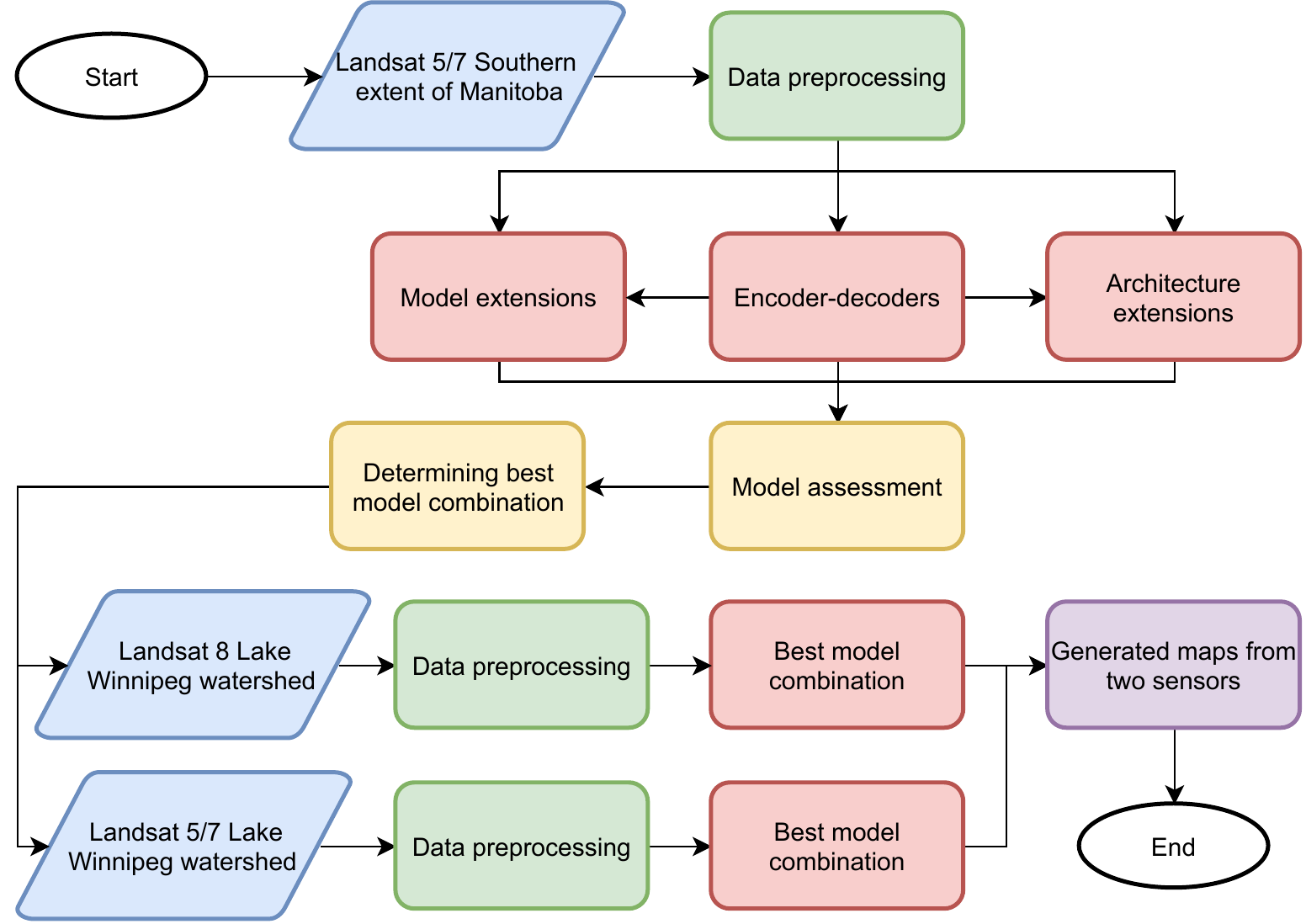}
	\caption[Training flowchart.]{Training flowchart, where the blue shapes correspond to data, green shapes correspond to preprocessing, red shapes correspond to models and extensions, yellow shapes correspond to model assessment, and purple shapes correspond to final generated maps.}
	\label{fig:process_flow}
\end{figure}

The experiments evaluated and compared the performance of different model variations, which were trained on the Landsat 5/7 southern Manitoba. Then, based on the results, the best model combination was selected to train models using Landsat 5/7 Prairie and Landsat 8 Lake Winnipeg watershed datasets. The model trained on the Landsat 5/7 dataset will allow us to generate LULC maps from 1984 to 2017, and the model trained on the Landsat 8 dataset will allow us to generate LULC maps starting from 2013 and onwards. Lastly, the results of all generated products are discussed, assessed, and compared to existing state-of-the-art models. The overall training process is depicted in Figure \ref{fig:process_flow}.

\subsection{Evaluation Metrics}
The experiment results were evaluated using OA, precision, recall, F1-score, and critical success index (CSI)\footnote{CSI is also known as a threat score, and identical to IoU} were used to evaluate the results. The OA is a metric that calculates the ratio between the number of correctly classified pixels and a total number of pixels in the image, thus providing a general assessment of the prediction, which can be easily compared and understood. Define $n_{ij}$ as the number of pixels in the image with ground truth label $i$ and corresponding predicted label $j$. Let $t_i=\sum_{j=1}^{C}{n_{ij}}$ denote the total number of pixels labelled with label $i$, $C$ is the number of classes, $n_{ii}$ are the number of pixels correctly predicted, and $n_{ji}$ are the number of incorrectly label pixels (with respect to label $i$).  \cite{VictorAutomatedLULC:2019}. The OA is defined as

\begin{equation}\label{EQ_Pixel_Accuracy}
    OA = \frac{\sum^{C}_{i=1}{n_{ii}}}{\sum^{C}_{i=1}{t_{i}}}.
\end{equation}

\begin{figure}
    \begin{subfigure}{0.46\textwidth}
        \centering
        \includegraphics[width=1\linewidth]{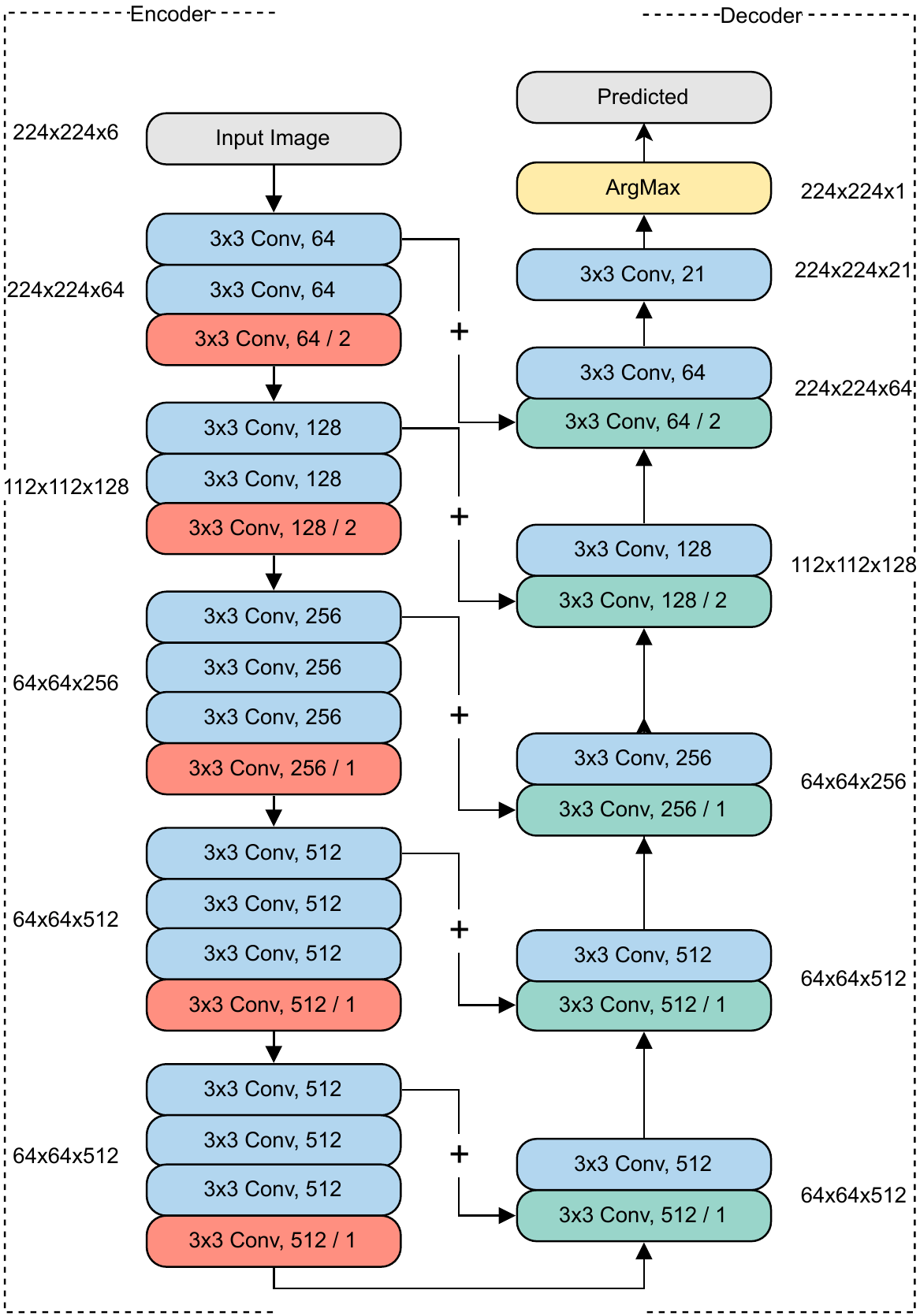}
        \caption[Structure of the best model]{}
        \label{fig:Best_model_1}
    \end{subfigure}
    \hfill
    \begin{subfigure}{0.43\textwidth}
        \centering
        \includegraphics[width=1\linewidth]{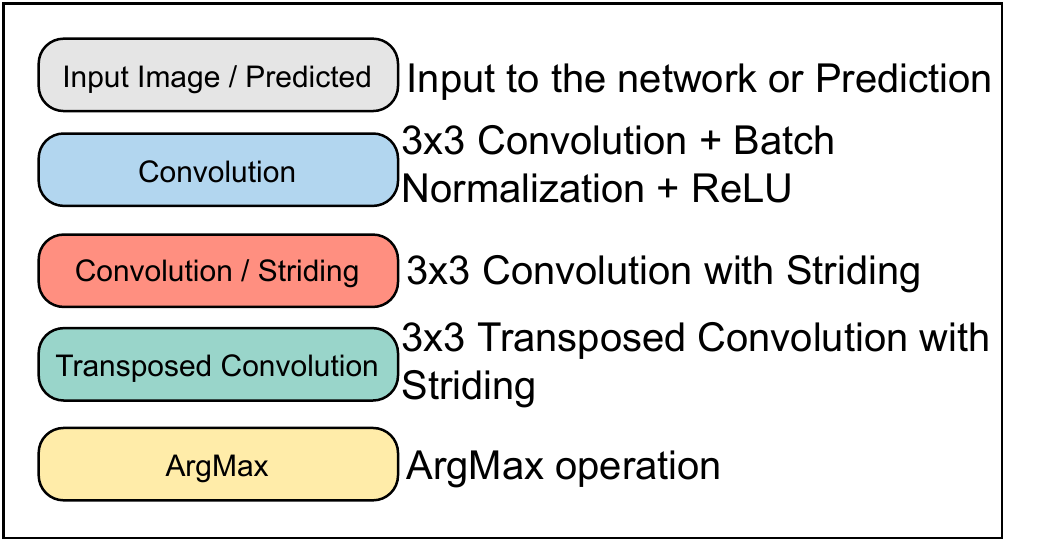}
        \caption[Description of each block]{}
        \label{fig:Best_model_2}
    \end{subfigure}
    \caption[Network architecture diagram for best model combination.]{Network architecture diagram of the structure for the best model combination: VGGNet encoder with modified U-Net decoder and OS 4 (a), and description of each block (b).}
    \label{fig:Best_model}
\end{figure}

Using the OA, the best model combination was determined, which was later trained on larger datasets to save time for the experiments. Next, we will define, precision \cite{Precision_Recall:1968}, recall \cite{Precision_Recall:1968}, F1-score \cite{FScore:1986}, and CSI \cite{CSI:1990}, where the CSI score is intersection over union (IoU) \cite{CSI_IOU:2003}. Let class $i$ be labelled a  \textit{positive class}, and every other class labelled a \textit{negative class}. Then, the true positive, $TP_i$, is defined as the number of correctly predicted pixels of the positive class, true negative $TN_i$  as the number of correctly predicted pixels of the negative class, false positive $FP_i$ as the number of incorrectly predicted pixels of the positive class, and false negative $FN_i$ as the number of incorrectly predicted pixels of the negative class. Given these definitions, we can then define the following metrics: 

\begin{subequations}

    \begin{equation} \label{EQ_Pixel_Precision}
        Precision_i = \frac{TP_i}{TP_i + FP_i};
    \end{equation}

    \begin{equation} \label{EQ_Pixel_Recall}
        Recall_i = \frac{TP_i}{TP_i + FN_i};
    \end{equation}

    \begin{equation} \label{EQ_Pixel_F1-Score}
        F1_i = 2 * \frac{Precision_i * Recall_i}{Precision_i + Recall_i};
    \end{equation}

    \begin{equation} \label{EQ_Pixel_CSI}
        CSI_i = \frac{TP_i}{TP_i + FP_i + FN_i}. 
    \end{equation}

\end{subequations}

Both the average and weighted average of each metric were calculated as

\begin{subequations}

\begin{equation} \label{EQ_Average}
    Average = \frac{1}{l}\sum^{l}_{i=1}M_i,
\end{equation}

\begin{equation} \label{EQ_WAverage}
    Weighted Average = \sum^{l}_{i=1}w_i*M_i. 
\end{equation}

\end{subequations}

where $l$ is the number of classes, $M_{i}$ is the metric value of the class $i$, and $w_i$ is the pixel proportion of the class $i$. Note, all the metrics were calculated on a per tile basis, and to get global dataset evaluation, each class $i$ was calculated using a weighted average.

\subsection{Comparison of Model Variations}
All model combinations were trained and validated on the southern Manitoba dataset. The experiment included the comparison of encoders, such as VGGNet, GoogleNet, ResNet, Xception, and decoders, such as FCN-8 \cite{FCN:2014}, U-Net, Feedbackward decoder \cite{FeedbackwardDecoding:2019}. Table \ref{tab_decoders_results} shows the results of the trained decoders paired with the VGGNet encoder. Based on the results, the modified U-Net decoder was chosen as the main decoder network, as it is a U-Net decoder with a reduced number of convolutions to one per upsampling layer. The chosen decoder has 3 times fewer parameters than plain U-Net with a cost of 0.04\% of OA. Table \ref{tab_encoders_results} displays the results of all implemented encoders in combination with the modified U-Net decoder. VGGNet proved to have the highest OA of 90.11\%, followed by ResNet at 89.05\%.

\begin{table}
    \centering
    \caption{Results of trained VGGNet encoder in combination with different decoders on the Landsat 5/7 data from the southern Manitoba.}
    \label{tab_decoders_results}
    \begin{tabular}{lc}
        \toprule
        Network         & OA      \\ 
        \midrule
        FCN-8 \cite{FCN:2014}           & 88.28                \\
        U-Net \cite{UNet:2015}           & \textbf{90.15}                \\
        Feedbackward \cite{FeedbackwardDecoding:2019}     & 89.59                \\
        Modified U-Net  & \textbf{90.11}                \\
        \bottomrule
    \end{tabular}
    
\end{table}

\begin{table}
    \centering
    \caption{Results of trained modified U-Net decoder in combination with different encoders on the Landsat 5/7 data from the southern Manitoba.}
    \label{tab_encoders_results}
    \begin{tabular}{lc}
    \toprule
        Network     & OA    \\
        \midrule
        VGGNet \cite{VGG:2015}       & \textbf{90.11}                \\
        GoogleNet \cite{InceptionV1:2014}    & 83.14                \\
        Xception \cite{Xception:2017}      & 88.36                \\
        ResNet \cite{ResnetV2:2016}      & \textbf{89.05}                \\
        \bottomrule
    \end{tabular}
\end{table}

\begin{figure}
\begin{subfigure}{0.49\textwidth}
    \begin{subfigure}{0.31\textwidth}
        \includegraphics[width=1\linewidth]{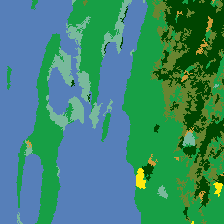}
    \end{subfigure}
    \begin{subfigure}{0.31\textwidth}
        \includegraphics[width=1\linewidth]{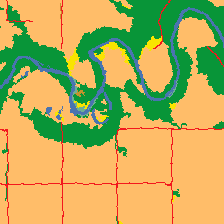}
    \end{subfigure}
    \begin{subfigure}{0.31\textwidth}
        \includegraphics[width=1\linewidth]{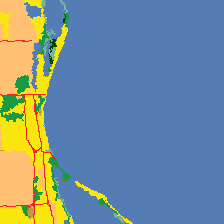}
    \end{subfigure}
    \vspace{0.15cm}

    \begin{subfigure}{0.31\textwidth}
        \includegraphics[width=1\linewidth]{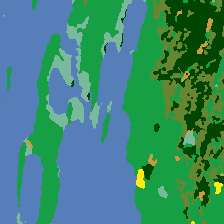}
    \end{subfigure}
    \begin{subfigure}{0.31\textwidth}
        \includegraphics[width=1\linewidth]{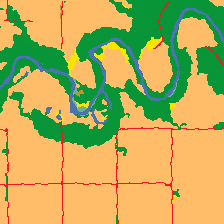}
    \end{subfigure}
    \begin{subfigure}{0.31\textwidth}
        \includegraphics[width=1\linewidth]{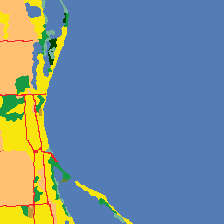}
    \end{subfigure}
    \caption{}
    \label{fig:Tiles_Landsat5_Small_Good_Results}
\end{subfigure}
\hfill
\begin{subfigure}{0.49\textwidth}
    \begin{subfigure}{0.31\textwidth}
        \includegraphics[width=1\linewidth]{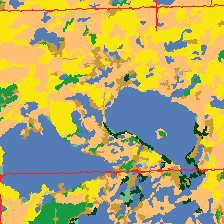}
    \end{subfigure}
    \begin{subfigure}{0.31\textwidth}
        \includegraphics[width=1\linewidth]{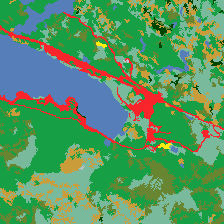}
    \end{subfigure}
    \begin{subfigure}{0.31\textwidth}
        \includegraphics[width=1\linewidth]{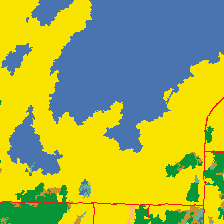}
    \end{subfigure}
    \vspace{0.15cm}
    
    \begin{subfigure}{0.31\textwidth}
        \includegraphics[width=1\linewidth]{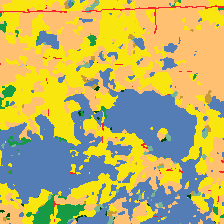}
    \end{subfigure}
    \begin{subfigure}{0.31\textwidth}
        \includegraphics[width=1\linewidth]{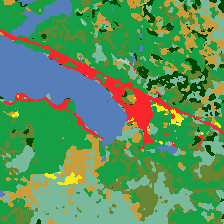}
    \end{subfigure}
    \begin{subfigure}{0.31\textwidth}
        \includegraphics[width=1\linewidth]{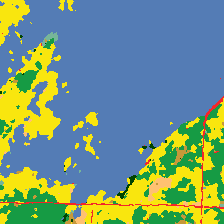}
    \end{subfigure}
\caption{}
\label{fig:Tiles_Landsat5_Small_Bad_Results}
\end{subfigure}

\caption[Examples of (a) good and (b) poor results generated on the Landsat 5/7 data from the southern Manitoba.]{Examples of (a) good and (b) poor results generated by the best model trained on the Landsat 5/7 data from the southern Manitoba. The first row are ground truth tiles, and second row are predicted tiles.}
    \label{fig:Tiles_Landsat5_Small_Results}

\end{figure}

VGGNet and ResNet CNNs with the modified U-Net decoder were chosen as backbone models for future experiments. During which both of these networks were trained with OS extension ranging from 32 to 2, where OS is 32 is the same as the models without extension. Additionally, networks were trained with different model and architectural extensions, such as ASPP, Context Module, GAN, Progressive GAN, and DeepLabv3+ (Table \ref{tab_Extensions}). Based on the results, adversarial training had lesser or no effect at all on the models that performed well on the given dataset. Also, the ResNet Deeplabv3+ architecture was trained on our dataset and showed low performance reaching 87.43\% of OA. The model with the best performance and lowest complexity was VGGNet with modified U-Net and OS 4, which reached 92.4\% on the southern extent of the Manitoba validation dataset. As the OA of the model approaches 100\% it becomes more difficult to improve the performance by introducing changes and enhancement to the model and architecture. In our experiments, we can see that extensions, like ASPP, context module and GAN can improve the performance of the model. However, if the model already performs well, implementing or even combining them will not provide further benefit, and in some cases even aggravate the results. The best performing network architecture is depicted in Figure \ref{fig:Best_model}. The best results generated by the model based on OA are shown in Figure \ref{fig:Tiles_Landsat5_Small_Good_Results}, where the model easily recognized edges of the water bodies, rivers and roads. Similarly, the worst results based on OA are shown in Figure \ref{fig:Tiles_Landsat5_Small_Bad_Results}, where the model did predict most of the area well but missed and misclassified some parts of the roads and water bodies.

\begin{table}
    \centering
	\caption{OA results of trained encoder-decoder models in combination with different extensions on the Landsat 5/7 data from the southern Manitoba.}
	\begin{adjustbox}{width=\columnwidth,center}
    \begin{tabular}{lcccccc}
    \toprule
    & \textbf{OS}&  32 & 16 & 8 & 4 & 2 \\
    \textbf{Network + Extension}  &&     &     &      &      &          \\
    \midrule
    VGGNet \cite{VGG:2015}                      && 90.11 & 90.66 & 91.25 & \textbf{92.4}  & 91.01     \\ 
    VGGNet + ASPP \cite{Deeplabv3+:2018}        && 90.72 & -     & -     & 92.22 & -         \\
    VGGNet + Context Module \cite{Context:2016}    && 90.18 & -     & -     & 92.04 & -         \\
    VGGNet + GAN \cite{SemanticGAN:2016}              && 89.99 & -     & -     & 92.37 & -         \\
    VGGNet + Progressive GAN \cite{ProgressiveGAN:2018,SemanticProgressiveGAN:2019}    && 90.09 & -     & -     & 92.41 & -         \\

    ResNet \cite{ResnetV2:2016}                     && 89.05 & 89.72 & 90.02 & \textbf{91.42} & -         \\
    ResNet + ASPP \cite{Deeplabv3+:2018}               && -     & -     & -     & 91.37 & -         \\
    ResNet + Context Module \cite{Context:2016}    && -     & -     & -     & 90.31 & -         \\
    ResNet + GAN \cite{SemanticGAN:2016}               && 89.26 & -     & -     & 91.04 & -         \\
    ResNet + Deeplabv3+ \cite{Deeplabv3+:2018}        && -     & -     & -     & 87.43 & -         \\

    \bottomrule
    \end{tabular}%
    \end{adjustbox}
    \label{tab_Extensions}%
\end{table}

\begin{table}
    \centering
	\caption{Pixel percentage of each class from the Landsat 5/7 Prairie and Landsat 8 Lake Winnipeg watershed datasets.}
	\label{tab_pixel percentage}
	\begin{adjustbox}{width=\columnwidth,center}
    \begin{tabular}{lcccc}
    \toprule
    &  \multicolumn{2}{c}{Pixel Percentage} &  \multicolumn{2}{c}{Number of Pixels} \\
    Name of the class & Landsat 5/7 & Landsat 8 & Landsat 5/7 & Landsat 8 \\
    \midrule
    No data                                           & 32.78 & 50.88    & 1031407726 & 1208979489  \\
    Temperate or sub-polar needleleaf forest          & 15.52 & 4.26     & 488262071 & 101124573  \\
    Sub-polar taiga needleleaf forest                 & 0.95  & 0.02     & 30020347 & 568110 \\
    Temperate or sub-polar broadleaf deciduous forest & 5.42  & 4.47     & 170429106 & 106208424  \\
    Mixed forest                                      & 3.06  & 2.21     & 96116915 & 52604376  \\
    Temperate or sub-polar shrubland                  & 5.84  & 1.52     & 183620158 & 36149337  \\
    Temperate or sub-polar grassland                  & 5.00  & 4.14     & 157337200 & 98413645  \\
    Sub-polar or polar shrubland-lichen-moss          & 0.13  & 1.80E-04 & 4117649 & 4270  \\
    Sub-polar or polar grassland-lichen-moss          & 0.35  & 1.70E-03 & 11029343 & 40467 \\
    Sub-polar or polar barren-lichen-moss             & 0.07  & 7.49E-06 & 2302774 & 178  \\
    Wetland                                           & 4.97  & 3.76     & 156313523 & 89369338  \\
    Cropland                                          & 13.75 & 20.27    & 432590674 & 481744136  \\
    Barren land                                       & 1.10  & 0.51     & 34526845 & 12135657  \\
    Urban and built-up                                & 0.85  & 1.12     & 26771188 & 26556386 \\
    Water                                             & 7.30  & 4.96     & 229553602 & 117838940\\
    Snow and ice                                      & 0.03  & 0.02     & 936557 & 534825  \\
    Cloud                                             & 2.89  & 1.85     & 90890767 & 43874765  \\
    \bottomrule
    \end{tabular}
    \end{adjustbox}
\end{table}

\begin{table}
    \centering
	\caption{Per-class assessment of the models trained on the Landsat 5/7 Prairie and Landsat 8 data from the Lake Winnipeg watershed.}
	\label{tab_per_class}
	\begin{adjustbox}{width=\columnwidth,center}
    \begin{tabular}{lcccc|cccc}
    \toprule
    & \multicolumn{4}{c}{Landsat 5/7}  & \multicolumn{4}{c}{Landsat 8} \\
    Name of the class & CSI & Precision & Recall & F1 & CSI & Precision & Recall & F1 \\
    \midrule
    No data                                           & 99.76 & 99.78 & 99.98 & 99.88 & 99.99 & 99.99 & 99.99 & 99.99 \\
    Temperate or sub-polar needleleaf forest          & 74.99 & 84.15 & 87.33 & 85.71 & 76.21 & 84.65 & 88.44 & 86.5  \\
    Sub-polar taiga needleleaf forest                 & 42.78 & 62.8  & 57.29 & 59.92 & 41.62 & 67.21 & 52.22 & 58.77 \\
    Temperate or sub-polar broadleaf forest           & 57.41 & 75.7  & 70.39 & 72.95 & 71.37 & 83.11 & 83.48 & 83.29 \\
    Mixed forest                                      & 40.78 & 63.27 & 53.42 & 57.93 & 56.03 & 74.1  & 69.67 & 71.82 \\
    Temperate or sub-polar shrubland                  & 49.67 & 65.29 & 67.5  & 66.38 & 48.61 & 70.48 & 61.04 & 65.42 \\
    Temperate or sub-polar grassland                  & 66.33 & 78.9  & 80.63 & 79.76 & 81.44 & 90.12 & 89.43 & 89.77 \\
    Sub-polar or polar shrubland-lichen-moss          & 21.68 & 55.47 & 26.24 & 35.63 & 17.62 & 54.06 & 20.73 & 29.97 \\
    Sub-polar or polar grassland-lichen-moss          & 42.36 & 57.97 & 61.13 & 59.51 & 21.25 & 59.3  & 24.87 & 35.04 \\
    Sub-polar or polar barren-lichen-moss             & 37.76 & 53.65 & 56.04 & 54.82 & 37.27 & 66.13 & 46.07 & 54.31 \\
    Wetland                                           & 58.08 & 73.75 & 73.21 & 73.48 & 75.76 & 85.78 & 86.64 & 86.21 \\
    Cropland                                          & 87.84 & 92.61 & 94.46 & 93.53 & 93.12 & 95.77 & 97.11 & 96.44 \\
    Barren land                                       & 72.2  & 82.01 & 85.79 & 83.85 & 70.79 & 84.13 & 81.7  & 82.9  \\
    Urban and built-up                                & 43.61 & 63.08 & 58.56 & 60.74 & 57.79 & 74.52 & 72.01 & 73.24 \\
    Water                                             & 86.36 & 94.45 & 90.98 & 92.68 & 88.25 & 95.04 & 92.51 & 93.76 \\
    Snow and ice                                      & 58.75 & 67.31 & 82.21 & 74.02 & 69.57 & 86.6  & 77.96 & 82.05 \\
    Cloud                                             & 97.58 & 97.76 & 99.81 & 98.78 & 91.51 & 94.81 & 96.33 & 95.56 \\
    \bottomrule
    \end{tabular}
    \end{adjustbox}
\end{table}

\subsection{Training Model on a Large Dataset}
\subsubsection{Landsat 5/7}
\begin{figure}
    \begin{subfigure}{0.49\textwidth}
        \begin{subfigure}{0.31\textwidth}
            \includegraphics[width=1\linewidth]{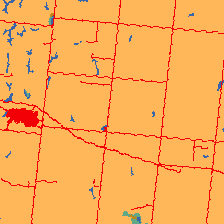}
        \end{subfigure}
        \begin{subfigure}{0.31\textwidth}
            \includegraphics[width=1\linewidth]{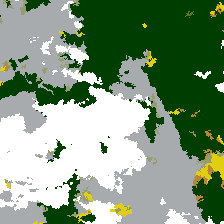}
        \end{subfigure}
        \begin{subfigure}{0.31\textwidth}
            \includegraphics[width=1\linewidth]{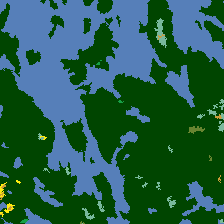}
        \end{subfigure}
        \vspace{0.15cm}

        \begin{subfigure}{0.31\textwidth}
            \includegraphics[width=1\linewidth]{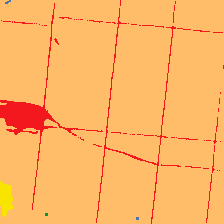}
        \end{subfigure}
        \begin{subfigure}{0.31\textwidth}
            \includegraphics[width=1\linewidth]{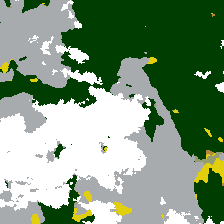}
        \end{subfigure}
        \begin{subfigure}{0.31\textwidth}
            \includegraphics[width=1\linewidth]{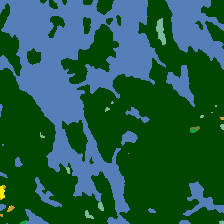}
        \end{subfigure}
        \caption{}
        \label{fig:Tiles_Landsat5_Big_Good_Results}
    \end{subfigure}
    \hfill
    \begin{subfigure}{0.49\textwidth}
        \begin{subfigure}{0.31\textwidth}
            \includegraphics[width=1\linewidth]{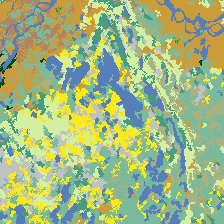}
        \end{subfigure}
        \begin{subfigure}{0.31\textwidth}
            \includegraphics[width=1\linewidth]{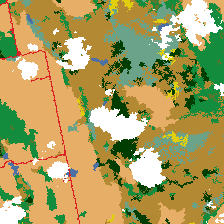}
        \end{subfigure}
        \begin{subfigure}{0.31\textwidth}
            \includegraphics[width=1\linewidth]{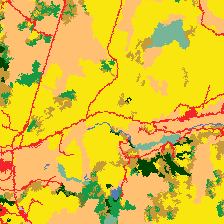}
        \end{subfigure}
        \vspace{0.15cm}
        
        \begin{subfigure}{0.31\textwidth}
            \includegraphics[width=1\linewidth]{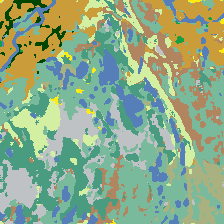}
        \end{subfigure}
        \begin{subfigure}{0.31\textwidth}
            \includegraphics[width=1\linewidth]{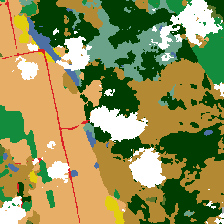}
        \end{subfigure}
        \begin{subfigure}{0.31\textwidth}
            \includegraphics[width=1\linewidth]{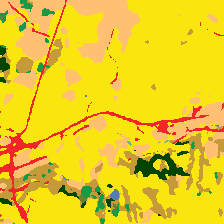}
        \end{subfigure}
    \caption{}
    \label{fig:Tiles_Landsat5_Big_Bad_Results}
    \end{subfigure}

    \caption{Examples of (a) good and (b) poor results generated by the best model trained on the Landsat 5/7 Prairie data. The first row are ground truth tiles, and second row are predicted tiles.}
        \label{fig:Tiles_Landsat5_Big_Results}
    
\end{figure}

After determining that the VGGNet encoder with modified U-Net decoder and OS 4 was the best, this model combination was then used to train models on the Landsat 5/7 Prairie dataset. Table \ref{tab_pixel percentage} (left) shows pixel proportion of each class, Table \ref{tab_per_class} (left) shows the per-class assessment of the Landsat 5/7 Prairie dataset and Table \ref{tab_best_model_datasets} shows a global assessment of the predicted map. Based on these results, LULC classes that performed well were found to have a higher percentage of pixels or representation than other LULC classes which had a small number of pixels. It has been documented that deep learning models need a large number of class examples (pixels) to achieve good performance for that class \cite{johnson2019survey}. Forest-related classes with low presence were the classes most often misclassified. Temperate or sub-polar needleleaf forest class and the urban and built-up classes were misclassified with cropland. Moreover, prediction of the cloud class had extremely high accuracy because the clouds were discriminated based on labels generated using a thresholding method \cite{Otsu:1979}; the network did not have difficulties generating this cloud class due to it being a straightforward function with sufficient data examples. Overall, the performance of the model reached 80.66\% OA on the validation dataset, where only a few tiles had no data pixels. No data is the background class that covers empty areas on the image. The accuracy presented in this dataset is lower due to the size and amount of extractable features presented in this dataset; particularly the discrimination of clouds. It was noted that the cloud labels generated by the Otsu thresholding method did not classify some of the clouds, especially transparent ones, and misclassified bright urban areas as clouds. This model showed the performance to be slightly better than published by \cite{latifovic:2016} but equivalent to the models present in ArcGIS/Pro deep learning package \cite{arcgis:2020}. The best results generated by the model based on OA are shown in Figure \ref{fig:Tiles_Landsat5_Big_Good_Results}, where the model had no difficulty in predicting tiles with a low number of classes, and perfectly segmented water, forests and croplands. Similarly, the worst results based on OA are shown in Figure \ref{fig:Tiles_Landsat5_Big_Bad_Results}, where the model missed some of the classes on the tiles with a high number of them, especially grassland, lichen-moss related classes and roads.

\subsubsection{Landsat 8}
Next, the same VGGNet encoder with modified U-Net decoder and OS 4 model was trained on the Landsat 8 Lake Winnipeg watershed which has higher radiometric resolutions (16-bit vs 8-bit of Landsat 5/7). The trend in the results provided by the model trained on the Landsat 8 dataset is similar to the Landsat 5/7 model and can be seen in Tables \ref{tab_per_class} and \ref{tab_best_model_datasets} (right). The OA of the predicted map increased by 8\%, and improvements in accuracy were noticed across most of the LULC classes. Since the land cover classes remained consistent, we speculate that these improvements are achieved due to the increased radiometric variability provided by the 16-bit instead of the 8-bit dataset, which made it easier to distinguish features due to a larger range of values. Recall, the cloud detection for Landsat 8 was trained using data generated from the Landsat 8 QA band provided by the sensor and not the thresholding method. Despite the results being lower than Landsat 7, the performance on the cloud class is still one of the best for this model. This model showed the performance higher than results reported by \cite{latifovic:2016} (11.44\% increase) and the models present in ArcGIS/Pro deep learning package \cite{arcgis:2020} by about ~6-8\%, and on the level of the state-of-the-art models designed for datasets with the lower amount of classes.

\begin{figure}
    \begin{subfigure}{0.49\textwidth}
        \begin{subfigure}{0.31\textwidth}
            \includegraphics[width=1\linewidth]{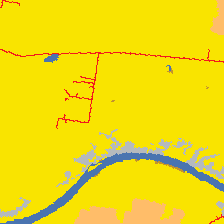}
        \end{subfigure}
        \begin{subfigure}{0.31\textwidth}
            \includegraphics[width=1\linewidth]{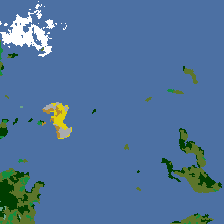}
        \end{subfigure}
        \begin{subfigure}{0.31\textwidth}
            \includegraphics[width=1\linewidth]{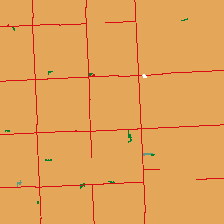}
        \end{subfigure}
        \vspace{0.15cm}

        \begin{subfigure}{0.31\textwidth}
            \includegraphics[width=1\linewidth]{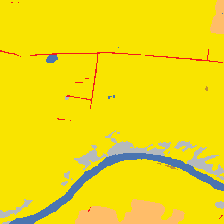}
        \end{subfigure}
        \begin{subfigure}{0.31\textwidth}
            \includegraphics[width=1\linewidth]{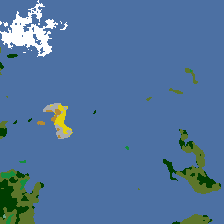}
        \end{subfigure}
        \begin{subfigure}{0.31\textwidth}
            \includegraphics[width=1\linewidth]{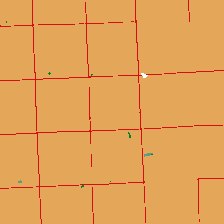}
        \end{subfigure}
        \caption{}
        \label{fig:Tiles_Landsat8_Big_Good_Results}
    \end{subfigure}
    \hfill
    \begin{subfigure}{0.49\textwidth}
        \begin{subfigure}{0.31\textwidth}
            \includegraphics[width=1\linewidth]{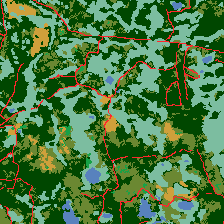}
        \end{subfigure}
        \begin{subfigure}{0.31\textwidth}
            \includegraphics[width=1\linewidth]{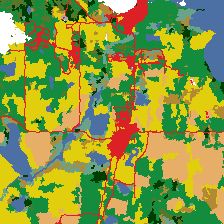}
        \end{subfigure}
        \begin{subfigure}{0.31\textwidth}
            \includegraphics[width=1\linewidth]{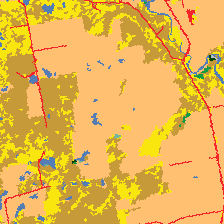}
        \end{subfigure}
        \vspace{0.15cm}
        
        \begin{subfigure}{0.31\textwidth}
            \includegraphics[width=1\linewidth]{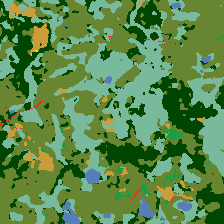}
        \end{subfigure}
        \begin{subfigure}{0.31\textwidth}
            \includegraphics[width=1\linewidth]{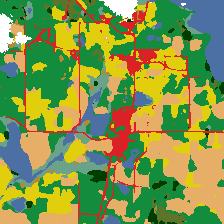}
        \end{subfigure}
        \begin{subfigure}{0.31\textwidth}
            \includegraphics[width=1\linewidth]{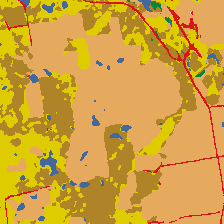}
        \end{subfigure}
    \caption{}
    \label{fig:Tiles_Landsat8_Big_Bad_Results}
    \end{subfigure}

    \caption{Examples of (a) good and (b) poor results generated by the best model trained on the Landsat 8 data from the Lake Winnipeg watershed. The first row are ground truth tiles, and second row are predicted tiles.}
        \label{fig:Tiles_Landsat8_Big_Results}
    
\end{figure}

\begin{table}
    \centering
	\caption{Global assessment of the models trained on the Landsat 5/7 Prairie and Landsat 8 data from the Lake Winnipeg watershed.}
	\label{tab_best_model_datasets}
    \begin{tabular}{lcc}
    \toprule
    Evaluation Metric & Landsat 5/7  & Landsat 8 \\
    \midrule
    OA          & 80.66 & 88.04 \\
        \midrule
    \multicolumn{3}{c}{Average} \\

    CSI         & 55.19 & 62.39 \\
    Precision   & 68.72 & 79.11 \\
    Recall      & 67.35 & 71.26 \\
    F1-Score    & 67.63 & 74.07 \\
        \midrule
    \multicolumn{3}{c}{Weighted Average} \\

    CSI         & 71.51 & 82.68 \\
    Precision   & 82.32 & 89.92 \\
    Recall      & 82.44 & 90.06 \\
    F1-Score    & 82.33 & 89.97 \\

    \bottomrule
    \end{tabular}
\end{table}

\newpage

\section{Conclusion}\label{Conclusion}
In this study, multiple NNs and extensions were trained on the southern Manitoba using the NALCMS dataset as labelled data. The results have shown that the best model variation is the VGGNet encoder with modified U-Net decoder and OS fixed at 4 that reached 92.4\% OA. Six different extensions were applied to the network, where most of them did not provide noticeable improvement both individually, and in combinations. Individual LULC class performance is highly dependent on the number of pixels representing that class; in general, the more pixels, the better the classification accuracy. Additionally, the best model variation was trained on the larger Landsat 5/7 Prairie datasets which resulted in a 80.66\% OA; an accuracy that is better than previously reported in the literature but equivalent to the performance of ArcGIS/Pro deep learning package \cite{arcgis:2020}. Also, the same model was trained on the Landsat 8 Lake Winnipeg dataset, where performance reached 88.04\% OA due to higher radiometric resolution. The Landsat 8 data increased the overall accuracy by 11.44\% compared to \cite{latifovic:2016} and is ~6-8\% higher than any other deep learning model present in the ArcGIS/PRo package that solves an equivalent problem on a similar satellite. And is on the level of the state-of-the-art DL models designed for datasets with a lower amount of classes. This shows that it is possible to produce fairly accurate NALCMS LULC maps from Landsat 5/7 and 8 sensors for 37 years of archived data and also future Landsat 8 data.

One of the limitations of DL is the amount of labelled data required to generated land cover map products. Thus, one of the most important directions of research in the future is to find a semi-supervised solution to produce accurate LULC maps, such as examination of unsupervised domain adaptation architectures \cite{UDAGAN:2017,UDA_GANRemoteSensing:2019}. Also, NNs can be trained on high-resolution satellite images to provide sharp and more detailed LULC maps. Furthermore, a multi-sensor model can be considered by training a network on multiple datasets from different sensors at once, making a more generic and versatile model, which can be used to develop LULC maps from the different sensors without prior training.


\bibliography{interactcadsample}   
\bibliographystyle{spiejour}   




\listoffigures
\listoftables

\end{document}